\documentclass[journal]{IEEEtran}
\usepackage{amsmath,amsfonts}
\usepackage{algorithmic}
\usepackage{array}
\usepackage[caption=false,font=normalsize,labelfont=sf,textfont=sf]{subfig}
\usepackage{textcomp}
\usepackage{stfloats}
\usepackage{url}
\usepackage{verbatim}
\usepackage{graphicx}
\usepackage{xcolor}
\usepackage{balance}

\renewcommand{\Re}{\mathbb{R}}

\def\BibTeX{{\rm B\kern-.05em{\sc i\kern-.025em b}\kern-.08em
    T\kern-.1667em\lower.7ex\hbox{E}\kern-.125emX}}

\begin{document}
\title{Development of Tendon-Driven Compliant Snake Robot with Global Bending and Twisting Actuation}

\author{Seongil Kwon, Serdar Incekara, Gangil Kwon, and Junhyoung Ha, \IEEEmembership{Member,~IEEE}
\thanks{This work was supported by the National Research Foundation (NRF) of Korea under the grant No. 2022R1C1C1005483 funded by the Korea government (MSIT). J.~Ha is affiliated with the Center for Healthcare Robotics, Department of AI Robot, Korea Institute of Science and Technology, Seoul, South Korea. {\it (Corresponding author: Junhyoung Ha.)}}
\thanks{S.~Kwon is with the Advanced Research Team, Endorobotics Co., Ltd., Seoul, South Korea (e-mail: ksi6123@gmail.com).}
\thanks{S.~Incekara and J.~Ha are with the Department of AI Robot, Korea Institute of Science and Technology, Seoul, South Korea (e-mail: serdarncekara@gmail.com;gangilgwon178@gmail.com;hjhdog1@gmail.com).}
\thanks{G.~Kwon is with the Department of FAB Engineering and Operations, Samsung Electronics, Suwon, South Korea (e-mail: gangilgwon178@gmail.com).}
\thanks{This work has been submitted to the IEEE for possible publication. Copyright may be transferred without notice, after which this version may no longer be accessible.}
}


\maketitle

\begin{abstract}
Snake robots have been studied for decades with the aim of achieving biological snakes' fluent locomotion. Yet, as of today, their locomotion remains far from that of the biological snakes. Our recent study suggested that snake locomotion utilizing partial ground contacts can be achieved with robots by using body compliance and lengthwise-globally applied body tensions. In this paper, we present the first hardware implementation of this locomotion principle. Our snake robot comprises serial tendon-driven continuum sections and is bent and twisted globally using tendons. We demonstrate how the tendons are actuated to achieve the ground contacts for forward and backward locomotion and sidewinding. The robot's capability to generate snake locomotion in various directions and its steerability were validated in a series of indoor experiments.
\end{abstract}


\section{Introduction}
\IEEEPARstart{S}{nake} robots have been a subject of research for several decades, driven by the desire to replicate the unique and efficient locomotion of biological snakes. The fascination with these robots stems from their potential applications in search and rescue missions, reconnaissance and exploration tasks, and industrial inspections, where their slender and flexible bodies can navigate through confined and complex environments that are inaccessible to conventional robots.

Early designs of snake robots have largely relied on rigid mechanisms~\cite{wright2007design, marvi2014sidewinding, transeth2008snake, transeth20083}. The famous modular snake robot developed by Carnegie Mellon University (CMU) demonstrated interesting motions including rod-climbing and sidewinding~\cite{wright2007design, wright2012design, rollinson2016pipe}. Similarly, rigid snake robots exhibited various motions by interacting with diverse environmental structures, such as pipes, holes, and obstacles~\cite{rollinson2016pipe, takemori2021hoop, transeth2008snake, travers2016shape, liljeback2011snake, liljeback2011experimental}. Tendon-driven actuation has also been used for snake robot development. A notable example is the tendon-driven snake robot presented in \cite{racioppo2019design}, which demonstrated multiple robot segments bent simultaneously using a single tendon. Recently, soft snake robots have been researched to overcome the lack of adaptability of rigid mechanisms. Similar to the rigid robots, they have shown various motions, facilitated by compliant interactions with environments~\cite{luo2014theoretical, branyan2017soft, branyan2020snake, liao2020soft, arachchige2023wheelless, rozaidi2023hissbot}.

The primary challenge of these robots lies in locomotion control. Biological snakes utilize the interactions with the ground and surrounding objects to propel. The obstacle-aided control methods have been developed for conventional robots in this regard~\cite{transeth2008snake, travers2016shape, liljeback2011snake, liljeback2011experimental}. However, their locomotion becomes challenging on a flat ground with no obstacles to interact with. In this scenario, sidewinding motions were mostly demonstrated with the existing snake robots~\cite{marvi2014sidewinding, transeth20083, arachchige2023wheelless, rozaidi2023hissbot, astley2015modulation}, or passive and active wheels were additionally utilized to facilitate forward motions toward the wheels' rolling direction~\cite{mori2002three, kamegawa2009realization, crespi2008online, fjerdingen2009snake, murugendran2009modeling}.

To achieve snake locomotion that closely resembles that of biological snakes, it is essential to align the mechanism design with the fundamental principles of snake movement. In environments without obstacles to push, e.g., on flat ground, biological snakes utilize their entire body to make certain patterns of partial ground contacts and compose the ground friction forces into a desired directional propulsion force~\cite{marvi2014sidewinding, astley2015modulation, hu2009mechanics, kano2020decoding}. In particular, unlike other types of biological locomotion, the snake's slithering motion (or the snake's forward undulatory locomotion) utilizes sliding friction to generate propulsion~\cite{hu2009mechanics}.
Considering that almost every biological locomotion utilizes static friction to establish stable bonding between their bodies and the ground (e.g., legged locomotion), this is such a unique locomotion in nature,
which poses a significant challenge for existing snake robots to control contact forces in their locomotion.
Nevertheless, these ground contact patterns and forces are seamlessly achieved by biological snakes even in rapid dynamic locomotion, which is in fact astonishing, considering they must 
control their body curvature and ground contact forces simultaneously and precisely to create the desired contacts and frictions varying in sync with their undulatory cycles.

Our recent study has highlighted that, although it was previously noted as important for navigating uneven terrains and obstacles, body compliance also plays a significant role in achieving ground contact forces required for various types of snake locomotion in environments without useful obstacles~\cite{ha2023robotic}. The core message of this study was that body compliance combined with body deflection under gravity naturally yields the required ground contact forces by applying lengthwise uniform body tensions in vertical bending and axial twisting. As this approach does not necessitate active contact force control, which has been the biggest barrier in conventional robots, it reduces the complexity of locomotion control. The primary challenge arising when applying this approach is the increased design complexity to enable adaptive control of the lengthwise uniform bending and twisting along the entire body, which we aimed to address in the present study.

In this paper, we present the first hardware implementation of the aforementioned locomotion principles. Our snake robot is composed of serial tendon-driven continuum sections that are compliance in bending and twisting. Each section can bend laterally to generate any planar motions. Furthermore, it is equipped with additional tendons routed globally through the entire robot length that can produce vertical bending and axial twisting along the entire body. This design enables the robot to exhibit forward and backward locomotion, as well as sidewinding,
by combining the planar undulation driven by the segments and the body tensions created by the globally routed tendons.

The remainder of this paper is organized as follows: Section~\ref{sec:related_works} reviews related works, providing a comprehensive overview of previous research and developments in the field of snake robots. Section~\ref{sec:method} outlines the methodology, including the hardware structure and control scheme for snake locomotion. Section~\ref{sec:hardware_details} details the component designs, the actuation method, and the mechatronics system for robot teleoperation. Section~\ref{sec:locomotion_control} describes the joint and motor control method, and Section~\ref{sec:experiments} presents the experimental results, followed by the conclusion in Section~\ref{sec:conclusion}.

\section{Related Works}
\label{sec:related_works}

Snake robots have been extensively studied to develop various mechanisms and control schemes. This section reviews key developments in both rigid and soft snake robots, as well as advances in locomotion control.

\subsection{Rigid Snake Robots}

Rigid snake robots can be categorized into wheeled and wheel-less types, as reviewed below.

\subsubsection{Wheeled Snake Robots}

Biological snakes utilize anisotropic skin properties to facilitate forward thrust, with minimal skin friction in the forward direction and greater friction in the backward and sideways directions~\cite{hu2009mechanics}. Wheels have been used in snake robots to replicate this frictional anisotropy. Passive wheels were incorporated in various designs~\cite{mori2002three, kamegawa2009realization, crespi2008online, yu2009amphibious, togawa2000study}, whereas active wheels were employed to provide greater control~\cite{fjerdingen2009snake, murugendran2009modeling}.

\subsubsection{Wheel-less Snake Robots}

Wheel-less snake robots were developed to overcome the typical limitation of wheeled robots, which is their restricted ability to navigate various terrains. These robots require more sophisticated control and planning, utilizing environmental contacts. One class of motion control for these robots is obstacle-aided control, where robots push or pull against surrounding objects to propel themselves~\cite{transeth2008snake, travers2016shape, liljeback2011snake, liljeback2011experimental, rollinson2014torque}.

Some robots have demonstrated motion in more complex environments. For example, the CMU modular snake robot and other similar robots have exhibited rod-climbing motion~\cite{rollinson2016pipe, wei2014structure, xiao2021adaptive}. Additionally, a robot capable of climbing a ladder was presented in \cite{takemori2018ladder}. The study in \cite{takemori2021hoop} introduced a non-sliding gait to navigate through virtually defined hoops. Snake locomotion inside pipes was also studied in \cite{rollinson2016pipe, trebuvna2016inspection, bando2016sound}.

\subsection{Soft Robots}

Soft snake robots utilize compliant materials and structures to achieve flexible and adaptive locomotion. These robots can deform their bodies to conform to various obstacles, providing a distinct advantage over rigid designs. A soft snake robot with passive wheels that exhibits pressure-operated undulation was developed in \cite{luo2014theoretical}. A novel soft actuator for undulatory locomotion and its gait patterns were presented in \cite{branyan2017soft}. Pneumatic-actuated soft robots demonstrating kirigami skin-aided lateral undulation and rod-climbing motion were presented in\cite{branyan2020snake} and~\cite{liao2020soft}, respectively. Sidewinding motions of soft snake robots were demonstrated in \cite{arachchige2023wheelless, rozaidi2023hissbot}.

\subsection{Locomotion Control}

Our review of locomotion control in snake robots focuses on wheel-less robots, which are more relevant to our study.

\subsubsection{Obstacle-Aided Control}

Many previous works on snake robot locomotion have focused on utilizing environmental objects to propel the robots. This approach is inspired by the behavior of biological snakes, which often use objects in their environment to push off and navigate through complex terrains. Obstacle-aided control mainly relies on lateral supports from objects to restrict sideways slip and only allow forward sliding~\cite{transeth2008snake, travers2016shape, liljeback2011snake, liljeback2011experimental}. These lateral supports provided by obstacles offer a similar advantage to using passive wheels, facilitating forward movement with minimal sideways slip.

\subsubsection{Undulatory Locomotion}

Biological snakes exhibit various types of undulatory locomotion, including sinus-lifting forward locomotion~\cite{hu2009mechanics}, center-lifting backward locomotion~\cite{kano2020decoding}, and sidewinding~\cite{marvi2014sidewinding, astley2015modulation}. These motions are based on similar body undulation, with different ground contacts between the motions causing different directional thrusts.

Sidewinding is the most commonly implemented locomotion mode in snake robots~\cite{marvi2014sidewinding, transeth20083, arachchige2023wheelless, rozaidi2023hissbot, astley2015modulation}. This is because sidewinding utilizes static ground contacts sequentially made over undulation cycles, which is fundamentally not different from legged locomotion. The physics of this locomotion mode is easier to understand and implement in robots.

Center-lifting backward locomotion is observed only in limited species~\cite{kano2020decoding} and has rarely been intentionally implemented in robots. In this motion, snakes make ground contacts at the left-most and right-most edges of the body curve. A similar motion principle can be found in pipe-navigating snake robots~\cite{rollinson2016pipe, trebuvna2016inspection, bando2016sound}. When these robots fit tightly in a pipe and undulate, they naturally make contact with the pipe at the side edges of their body curve.

Although sinus-lifting forward locomotion is the most representative snake locomotion, it has rarely been demonstrated in snake robots. This motion relies on sliding contacts, which are difficult to model and control in robots. To the best of the authors' knowledge, our previous study~\cite{ha2023robotic} was the only work that demonstrated sinus-lifting forward locomotion using a wheel-less snake robot. The use of body compliance and ground contacts caused by natural body deflection enabled this motion, as well as center-lifting backward locomotion and sidewinding.

\section{Method}
\label{sec:method}

In this section, we first review the underlying principles of snake locomotion as discovered in our previous work~\cite{ha2023robotic}, followed by an overview of our hardware implementation to realize these locomotion principles in a robotic snake mechanism.

\subsection{Locomotion Principle}
\label{subsec:principles}

Our previous study on snake locomotion revealed that body compliance and lengthwise-globally applied vertical bending and axial twisting can enable natural snake movement in snake robots~\cite{ha2023robotic}. Specifically, the following principles can be used to achieve undulatory snake locomotion:
\begin{itemize}
    \item Upward vertical bending during planar undulation results in sinus-lifting forward motion;
    \item Downward vertical bending during planar undulation results in center-lifting backward motion;
    \item Axial twisting during planar undulation results in sidewinding.
\end{itemize}

\begin{figure}[t]
	\subfloat[]{
		\includegraphics[width=0.99\columnwidth]{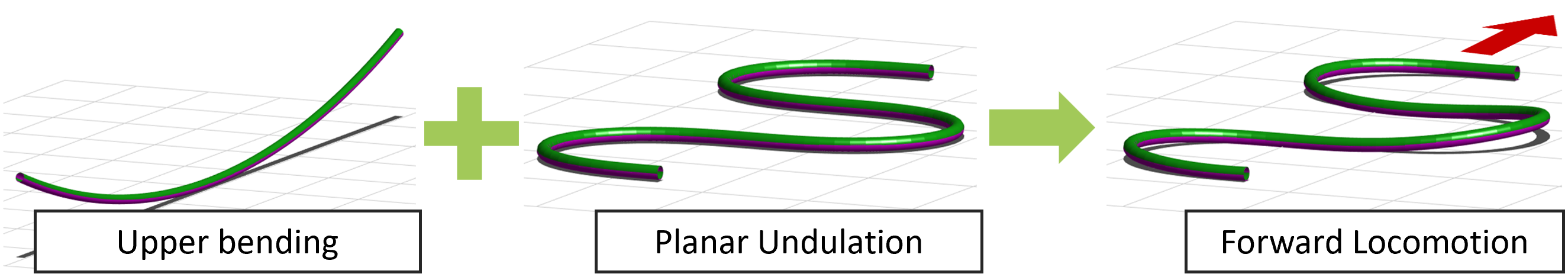}
	}
	\\
	\subfloat[]{
		\includegraphics[width=0.99\columnwidth]{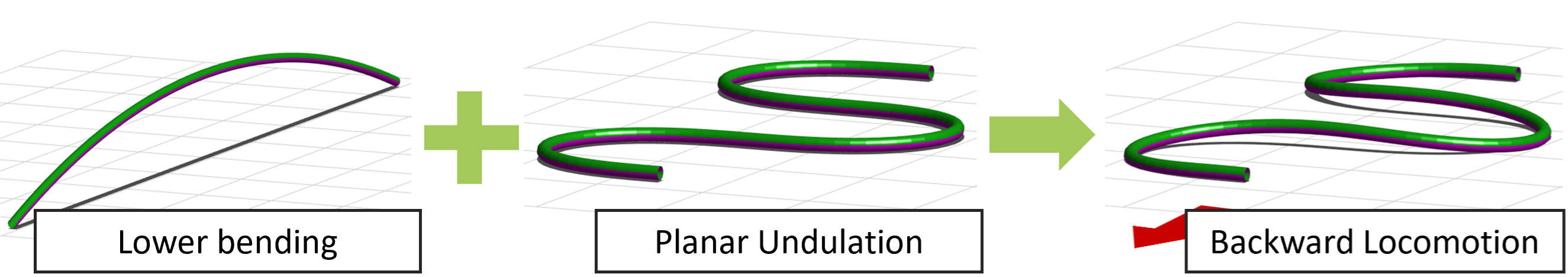}
	}
	\\
	\subfloat[]{
		\includegraphics[width=0.99\columnwidth]{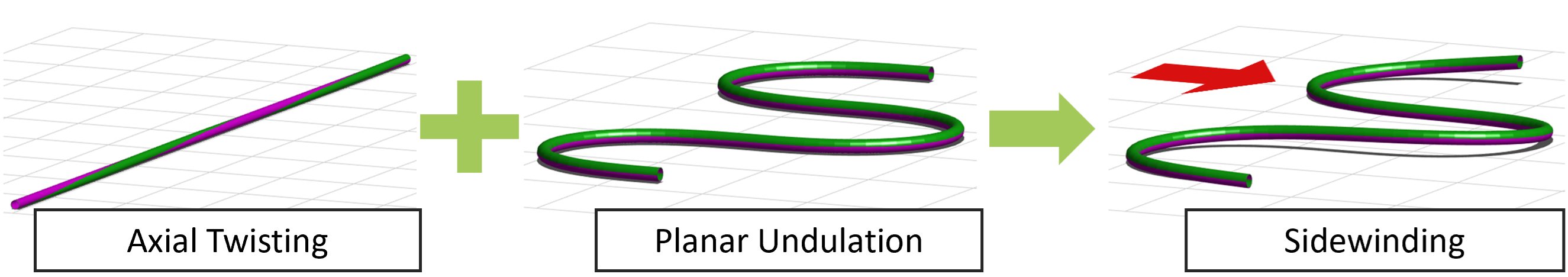}
	}
\caption{Undulatory locomotion principles using lengthwise-globally applied body tensions:  (a) forward locomotion, (b) backward locomotion, and (c) sidewinding. The red arrows represent the locomotion directions, and the body is bi-colored to visualize the body twist.}
\label{fig:locomotion_control}
\end{figure}

\subsection{Implementation Overview}

To apply our locomotion principles described in Section \ref{subsec:principles}, the robot hardware must be capable of adaptively controlling vertical bending and axial twisting during motion. This poses a significant challenge to the design of the robot mechanism. Herein, we overview how we have implemented these actions in our tendon-driven snake robot mechanism. More details on the module and joint design, as well as the mechatronics system, are provided in the following section.

\subsubsection{Hardware Configuration}

\begin{figure}[!t]
    \centering
    \includegraphics[width=3.5in]{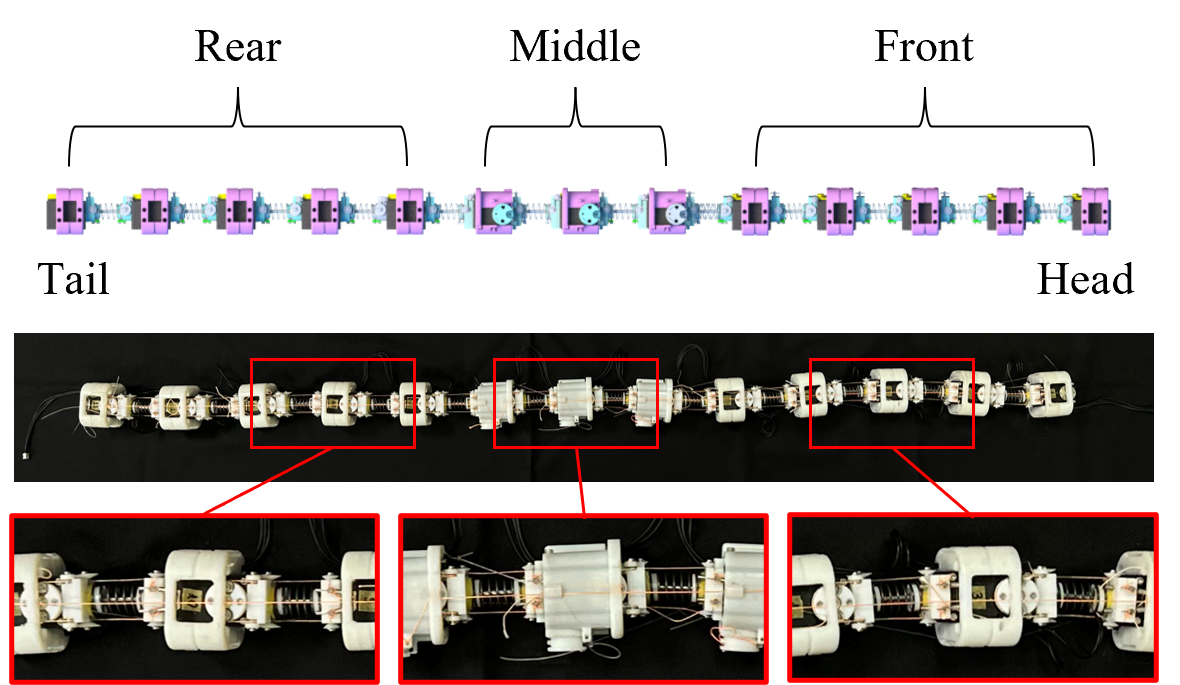}
    \caption{Our snake robot mechanism consists of $13$ motors arranged in series, with the three central motors being dual-axis. These dual-axis motors utilize their second axes for vertical bending and axial twisting.
    }
    \label{fig:straight_configuration}
\end{figure}

Our robot is composed of $13$ modules in series, each module equipped with a servo motor. We utilized ten single-axis and three dual-axis servo motors, with five single-axis motors at the front and five at the rear, and the three dual-axis motors in the middle, as depicted in Fig.~\ref{fig:straight_configuration}. Continuum joints are implemented between the motors, each comprising a spring and a central backbone, to provide compliance to the robot. Tendons are used to bend the continuum joints driven by adjacent motors. The overall weight and length were $2.21$ kg and $1.25$ m, respectively.

The two key elements in implementing our locomotion principles are planar undulation and vertical bending/axial twisting actuation. For planar undulation, the $12$ continuum joints alternate bending left and right at different phases between motors, producing a propagation of flexural waves along the robot's length. Additionally, vertical bending and axial twisting are actuated by the second axes of the dual-axis motors: the first axes are used for planar undulation, while the second axes are connected to additional tendons routed globally through the robot's body--straight paths along the back (top surface) and abdomen (bottom surface) and spiral paths around the body. This configuration allows for lateral bending of the continuum joints to create undulation, with globally routed tendons controlling ground contacts during the movement.

\subsubsection{Global Bending and Twisting Actuation}

\begin{figure}[!t]
    \centering
    \includegraphics[width=3.5in]{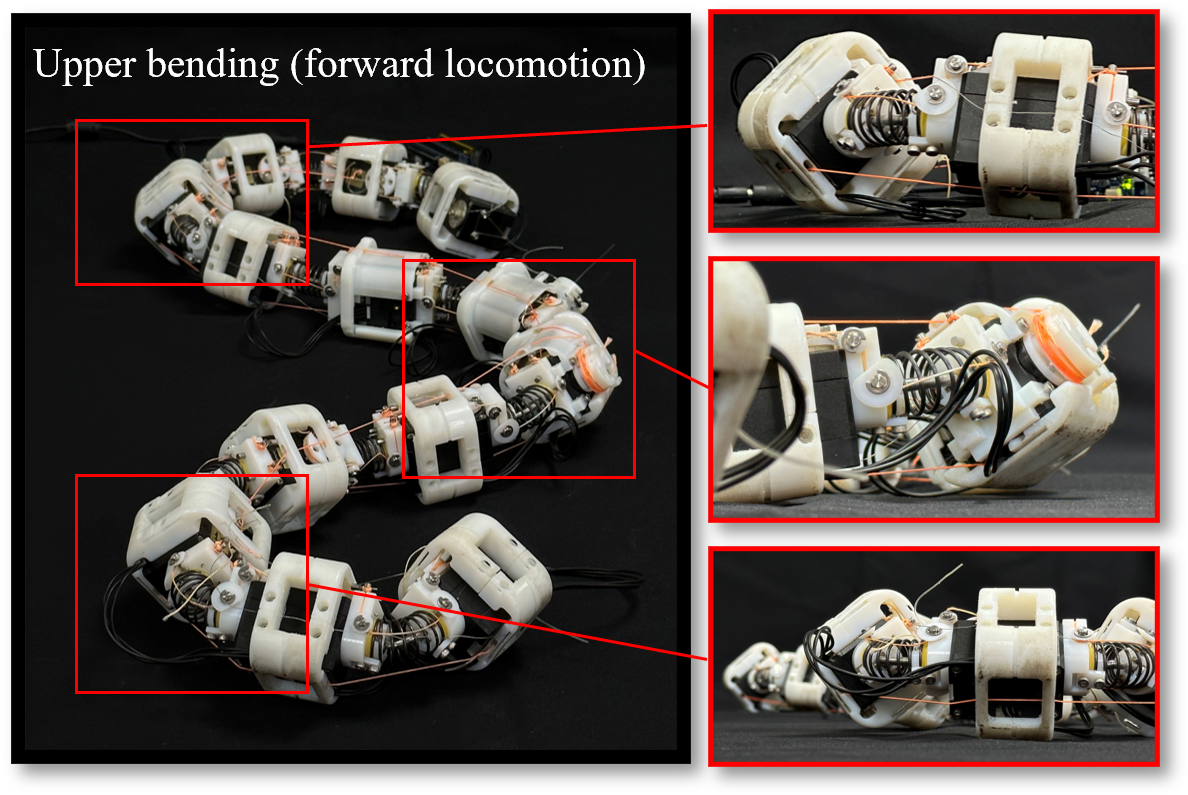}
    \caption{
    Robot configuration with vertical upper bending: In this configuration, the ground contacts facilitate sinus-lifting forward locomotion.}
    \label{fig:forward}
\end{figure}

Fig.~\ref{fig:forward} shows an S-shaped configuration of our snake robot with upper vertical bending achieved by pulling the globally routed tendons that pass through the back of the robot. The smaller images in Fig.~\ref{fig:forward} highlight the zoomed-in parts of the robot corresponding to the red boxes, where ground lifting occurs. This actuation is intended to generate forward locomotion, as inferred from the resulting robot configuration that resembles the body curvature of biological snakes in their sinus-lifting forward motion.

\begin{figure}[!t]
    \centering
    \includegraphics[width=3.5in]{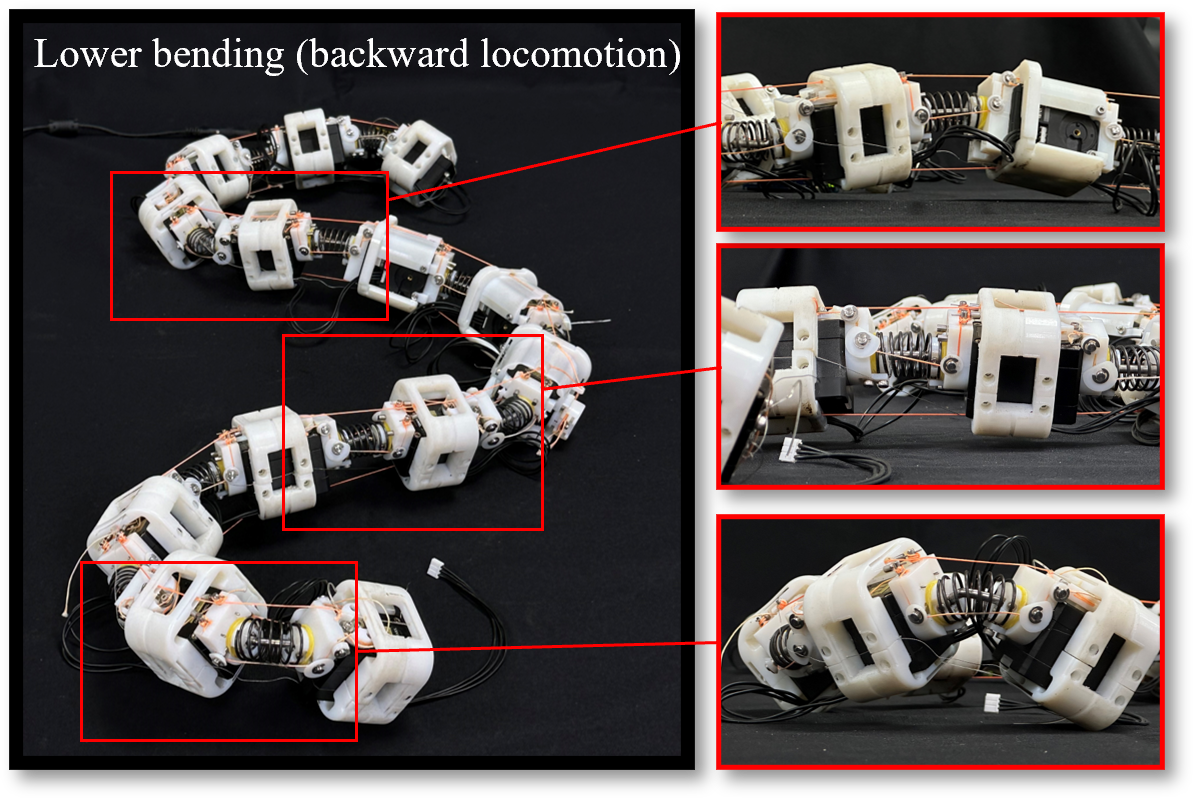}
    \caption{
    Robot configuration with vertical lower bending: In this configuration, the ground contacts facilitate center-lifting backward locomotion.}
    \label{fig:backward}
\end{figure}

Similarly, Fig.~\ref{fig:backward} demonstrates an S-shaped configuration of our robot with downward vertical bending achieved by pulling the globally routed tendons along the abdomen. The red boxes indicate the parts of the robot that lift off the ground, with zoomed-in images provided on the right side in Fig.~\ref{fig:backward}. This configuration corresponds to backward locomotion, which was observed in the center-lifting motion of Cerastes vipera~\cite{kano2020decoding}. In contrast to forward locomotion, the robot maintains ground contact at the leftmost and rightmost edges of its body, while the central sections are elevated.

\begin{figure}[!t]
    \centering
    \includegraphics[width=3.5in]{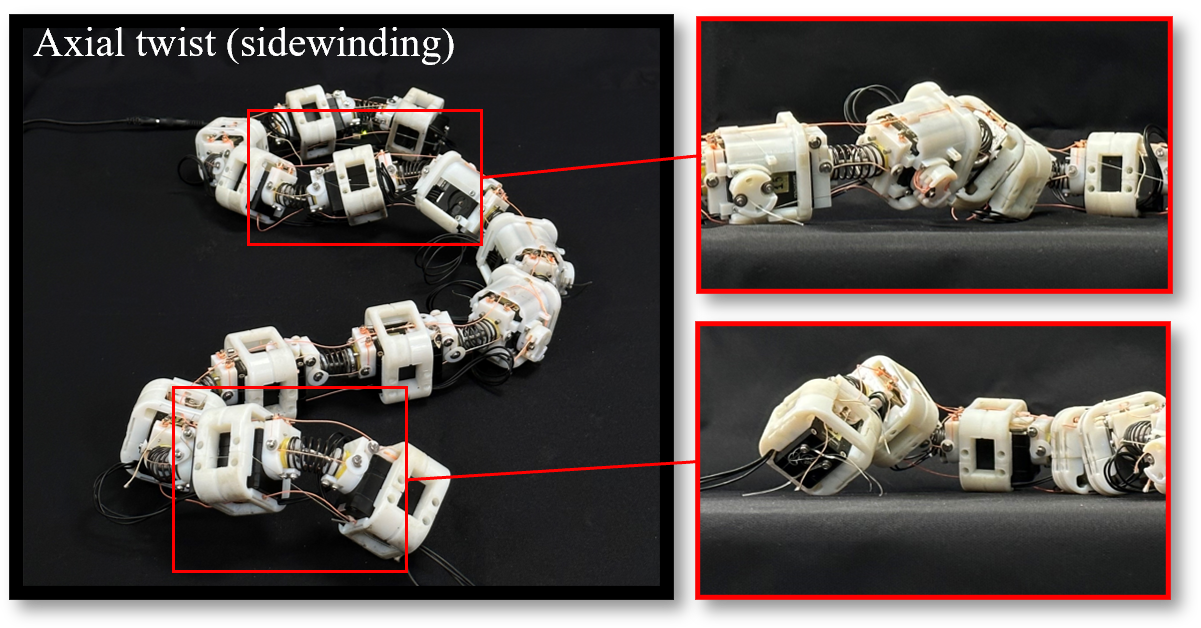}
    \caption{
    Robot configuration with axial twisting: In this configuration, the ground contacts facilitate sidewinding.}
    \label{fig:side}
\end{figure}

Lastly, the robot configuration under the tension of the spiral tendons is shown in Fig.~\ref{fig:side}. When the tendons are pulled, the robot twists in the opposite direction of the spiral tendon paths. When this actuation is combined with planar undulation, the sections of the robot body between the leftmost and rightmost edges alternately contact and lift off the ground, as indicated by the elevations in the red boxes in the figure. The smaller images on the right side are zoomed-in views of the corresponding red boxes. This ground contact pattern resembles that of biological snakes during sidewinding, where the ground contacts form diagonal lines~\cite{marvi2014sidewinding, astley2015modulation}.


\section{Hardware Development Details}
\label{sec:hardware_details}

This section presents more detailed descriptions of the hardware development and mechatronics system, including the continuum joint with its actuation method, the module design, the global bending and twisting actuation, and the overall mechatronics system for robot teleoperation.

\subsection{Continuum Joint}

\begin{figure}[!t]
\centering
\includegraphics[width=3.54in]{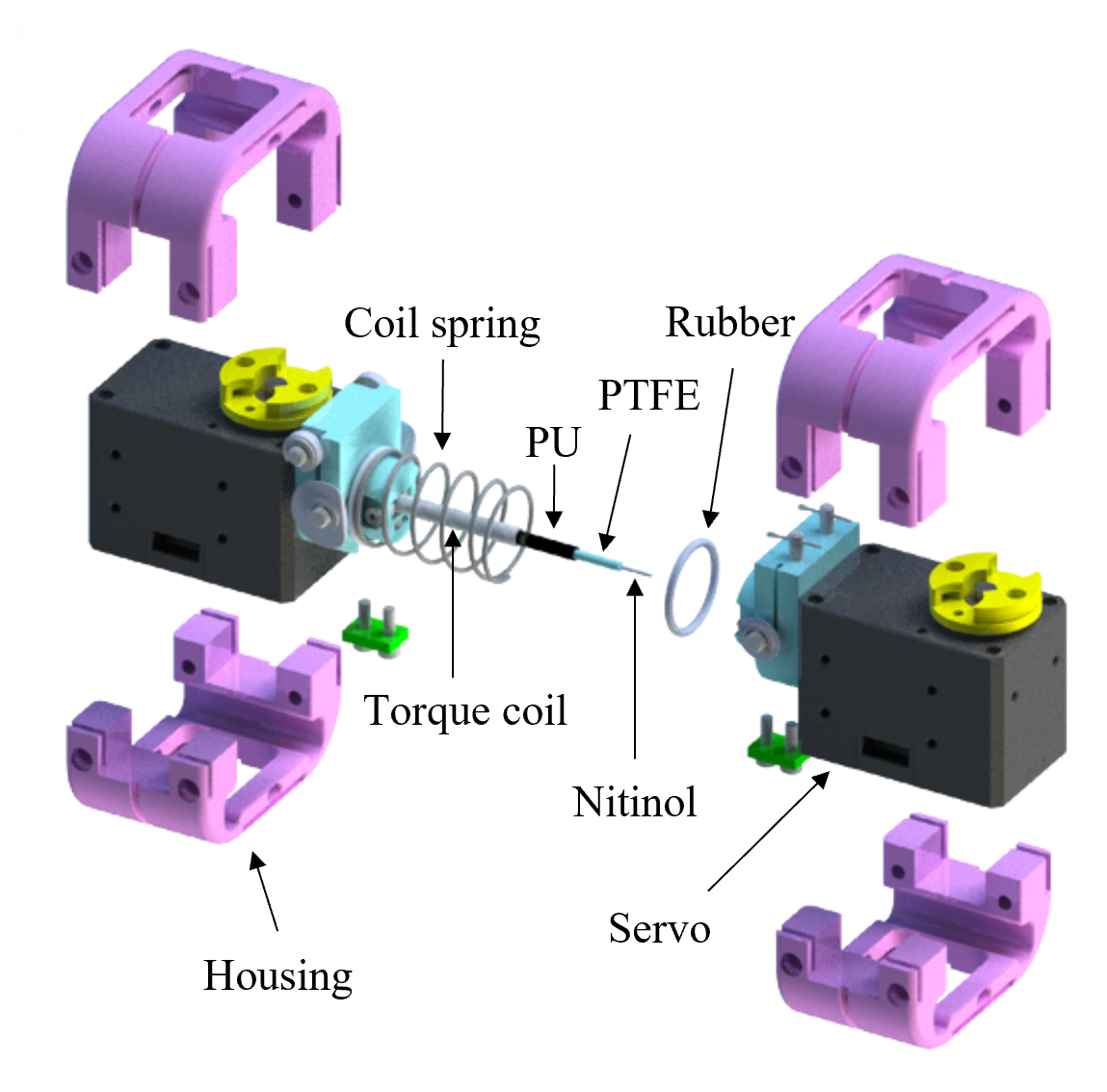}
\caption{Components of our continuum joint and adjacent modules. The rubber is used to fill the gap between the coil spring and the adaptor.}
\label{fig:joint}
\end{figure}

Our locomotion principles in \cite{ha2023robotic} rely on the robot's natural deflection caused by the robot's body compliance, not only in bending but also in torsion. To provide compliance to the robot, we developed continuum joints to connect between modules, each of which was built with a coil spring and a central backbone, as illustrated in Fig.~\ref{fig:joint}.

The coil springs possess a restorative force that allows them to smoothly bend in response to external forces and return to their original position when the external force is released, which gives bending compliance to the robot mechanism. To assemble the springs with the adjacent modules, $3$D printed adaptors were attached to the motors, and the springs were tightly fixed to the adaptors (see Fig.~\ref{fig:joint}). The springs bend through tendons connected between consecutive modules, which will be described in more detail in the following subsection.

While the tendon tensions cause bending in the springs, they can also lead to spring compression. This means that not all tension is used for pure bending, which reduces the bending amplitude of the joints during planar undulation. To minimize this bending loss, a central backbone was placed at the center of each spring, preventing spring compression while allowing for bending.

Although our locomotion principles require torsional compliance, high flexibility in torsion can cause unpredictable robot behaviors, such as tangling and rolling. While the coil spring is relatively stiffer in torsion than in bending compared to other solid structures (e.g., rods), we encountered issues when the springs were used alone for torsional resistance: the tension from multiple tendons during planar undulation could easily cause torsion, disrupting stable ground contact and leading to the robot flipping over. To mitigate this behavior, we included a torque coil in the central backbone of each joint. The torque coil exhibits flexible bending but excellent torque transmission, making it ideal for this application. Typically, it is made by winding two stainless steel wires in opposite spiral directions, resulting in low bending stiffness and high torsional stiffness. We used a torque coil manufactured by Asahi Intecc, with an inner diameter of $3$ mm and an outer diameter of $3.9$ mm. To prevent buckling, we filled the inside of the torque coil with a NiTi rod with a diameter of $0.8$ mm, along with PTFE and PU tubes.


\subsection{Joint Actuation Mechanism}

\begin{figure}[!t]
    \centering
    \subfloat[]{\includegraphics[width=1.25in]{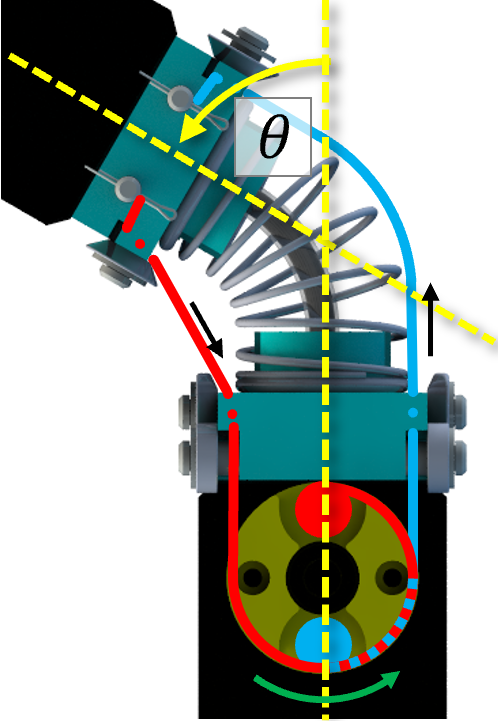}}
    \subfloat[]{\includegraphics[width=1.75in]{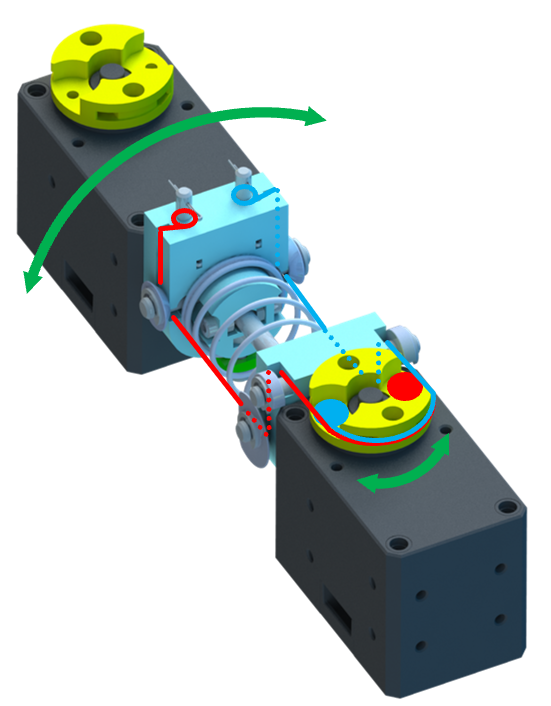}}
    \caption{Joint actuation mechanism: (a) Joint bending induced by tendon displacement.
    (b) Tendon paths with multiple $90^\circ$ turns.}
    \label{fig:joint_actuation}
\end{figure}

The continuum joints are attached to consecutive motors using the $3$D-printed adaptors. As depicted in Fig.~\ref{fig:joint_actuation}, two tendons are connected from the rotor of one motor to the adaptor of the adjacent motor. As the motor rotates, one tendon is pulled while the other is released, causing the continuum joint to bend toward the pulled tendon. This bending motion produces the planar undulation of the robot.

The tendon routing is of great importance in the design of the actuation mechanism. In particular, it  is crucial that the tendons run parallel to the central backbone of the joint on the same horizontal plane to ensure that the joints bend only in the left and right directions without tilting upwards or downwards. Since the ends of the tendons are located at the top of the motors, directly connecting the tendons in the shortest paths results in horizontal misalignments between the tendons and the backbone. To resolve this issue, slits were made in the adapters to allow the tendons to pass through. As shown in Fig.~\ref{fig:joint_actuation}(b), the tendons run through multiple $90^\circ$ turns via stainless steel pillars attached to the adaptors. The surfaces of the pillars were confirmed to be smooth to minimize friction when the tendons slide over them. To prevent the tendons from deviating from their paths during motor rotation, circular guides are attached to both sides of each pillar.

\begin{figure*}[t!]
\centering
\subfloat[]{\includegraphics[width=2.5in]{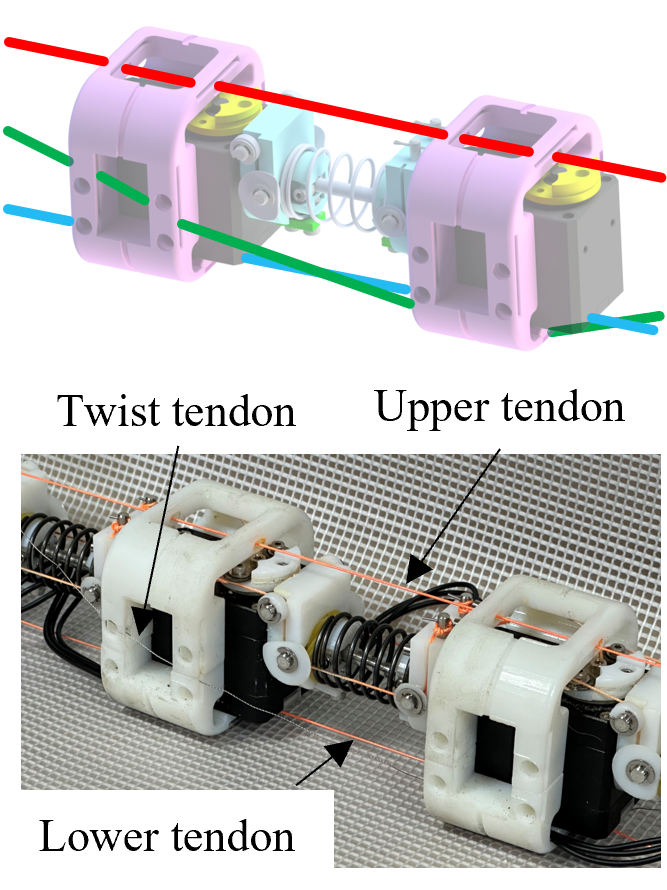}}
\hspace{0.5in}
\subfloat[]{\includegraphics[width=3.5in]{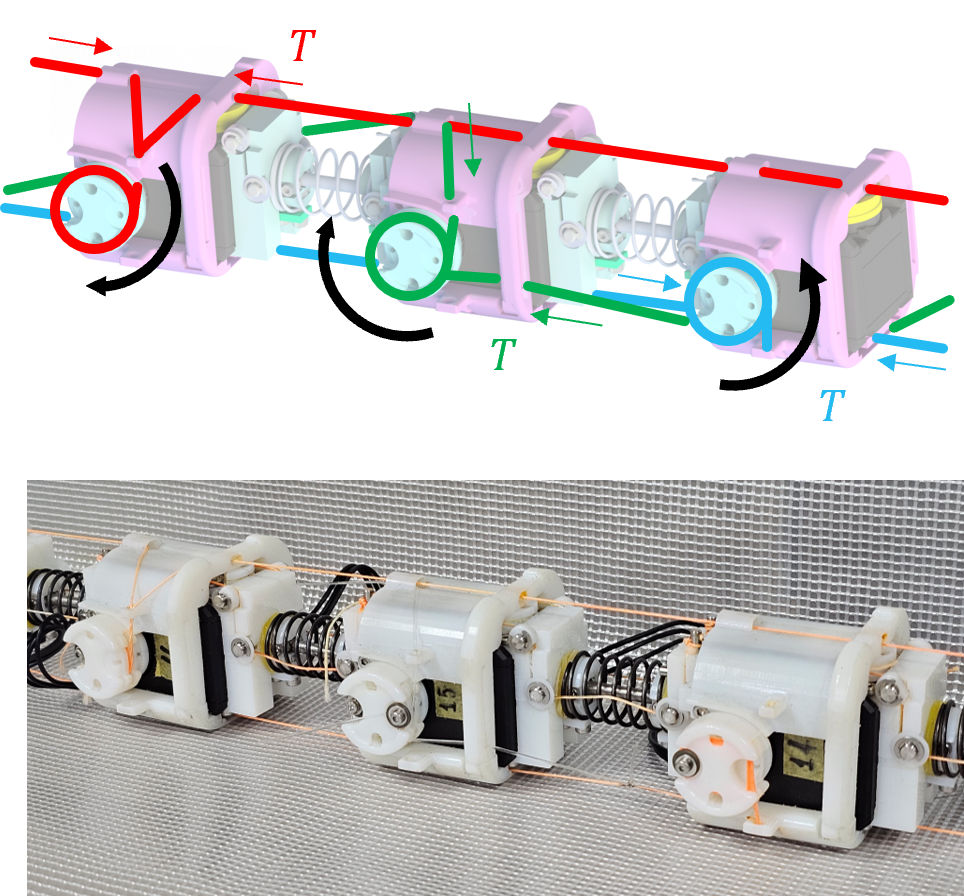}}
\\
\centering
\subfloat[]{\includegraphics[width=7in]{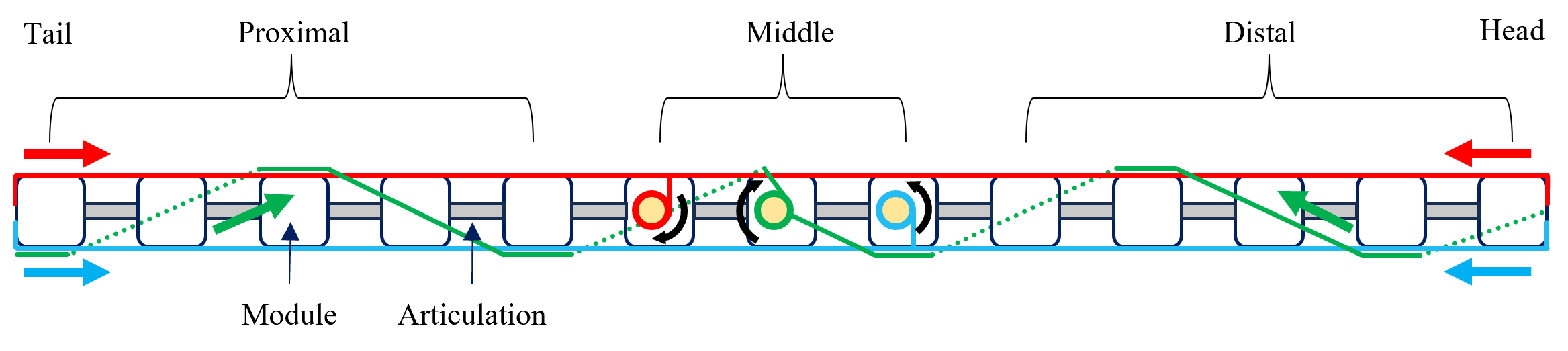}}
\caption{Diagrams and physical hardware images illustrating the paths of globally routed tendons:
(a) Tendon paths on modules with single-axis motors. (b) Tendon paths on modules with dual-axis motors. (c) Tendon paths over the entire robot length.
The red tendons are responsible for the upper bending required for forward locomotion, while the blue tendons enable the lower bending needed for backward locomotion. Additionally, the green tendons are routed through the housings in spiral paths to create an axial twist for sidewinding.}
\label{fig:assembly_module}
\end{figure*}

\subsection{Module Housing and Global Bending/Twisting Actuation}

As illustrated in Fig.~\ref{fig:joint}, each module is wrapped with a $3$D-printed plastic housing to provide the robot with enhanced durability and tendon paths for the global bending and twisting actuation. We fabricated the housing using VeroPureWhite (Stratasys Ltd., USA), known for its high strength and printing precision, making it suitable for our purpose.

The actuation mechanism for global bending and twisting involves tendons running through the entire length of the robot in straight and spiral paths, as depicted in Fig.~\ref{fig:assembly_module}. A key aspect of this system is the use of dual-axis motors. The first axes of these motors are used for the planar bending actuation of the continuum joints. As shown in Fig.~\ref{fig:assembly_module}(b), the second axes are located on the side of the motors. We designed rotors for these axes, where the globally routed tendons are connected. The other ends of the tendons are fixed at the head and tail of the robot. By rotating the motors over multiple revolutions, the tendons wrap around the rotors, resulting in tendon tensions for vertical bending and axial twisting. These actions are taken differently in different locomotion types. For forward movement, the red tendons in Fig.~\ref{fig:assembly_module} are wound while the blue tendons are unwound, whereas, in backward locomotion, the tendons are wound or unwound in the opposite way. In sidewinding, only the green cables are wound.

The dual-axis motors are placed in the middle of the robot (i.e., 6th, 7th, and 8th modules from either the head or the tail). Consequently, each second axis of the dual-axis motors is connected to two tendons, one going to the head and the other to the tail. This motor placement was determined based on our experimental results, which showed significant tension loss due to friction at the tail when the dual-axis motors were located at the head. By placing the motors in the middle, the tension losses are symmetrically distributed over the length of the robot.

Additionally, it is worth noting that the tendons used for joint bending are highly rigid in elongation as compliance naturally comes from the joint structure. However, using rigid tendons for global bending and twisting caused issues with inflexible motion due to the tendon length constraints. To address this, we concatenated flexible urethane cables with rigid cables, attaching the flexible ends to the head and tail and the rigid ends to the motors. This combination provides greater compliance in vertical bending and axial twisting, enabling more fluent snake locomotion.

\subsection{Mechatronics System for Actuation and Teleoperation}

Potential issues with using multiple motors in series include complicated wiring for signal and power and the space occupied by motor drivers. To avoid these issues, we used Dynamixel motors, manufactured by ROBOTIS Inc., Korea, which are equipped with built-in drivers and support the TTL communication protocol. The TTL communication enables serial connection of the motors, which significantly simplifies power and signal wiring.

Specifically, for single-axis motors, the XH430-W350-T model was chosen, and for dual-axis motors, the 2XC430-W250-T model was selected. These motors can generate $3.4$ Nm and $1.4$ Nm of torque, respectively. Their dimensions (w$\times$h$\times$d) are 28.5 mm $\times$46.5 mm$\times$34 mm and 36 mm$\times$46.5 mm$\times$36 mm, and they weigh 82 g and 98 g, respectively.

The user interface includes a desktop application and a wireless joystick. To communicate with the motors, we used an Arduino MKR WiFi 1010 and its compatible Dynamixel Shield. The Arduino board reads motor values and sends commands to the motors through the Dynamixel Shield. The desktop application was developed to monitor the status of the motors and send motion parameters and joystick inputs to the Arduino board via Bluetooth Low Energy (BLE).

\begin{figure}[!t]
\centering
\includegraphics[width=3.54in]{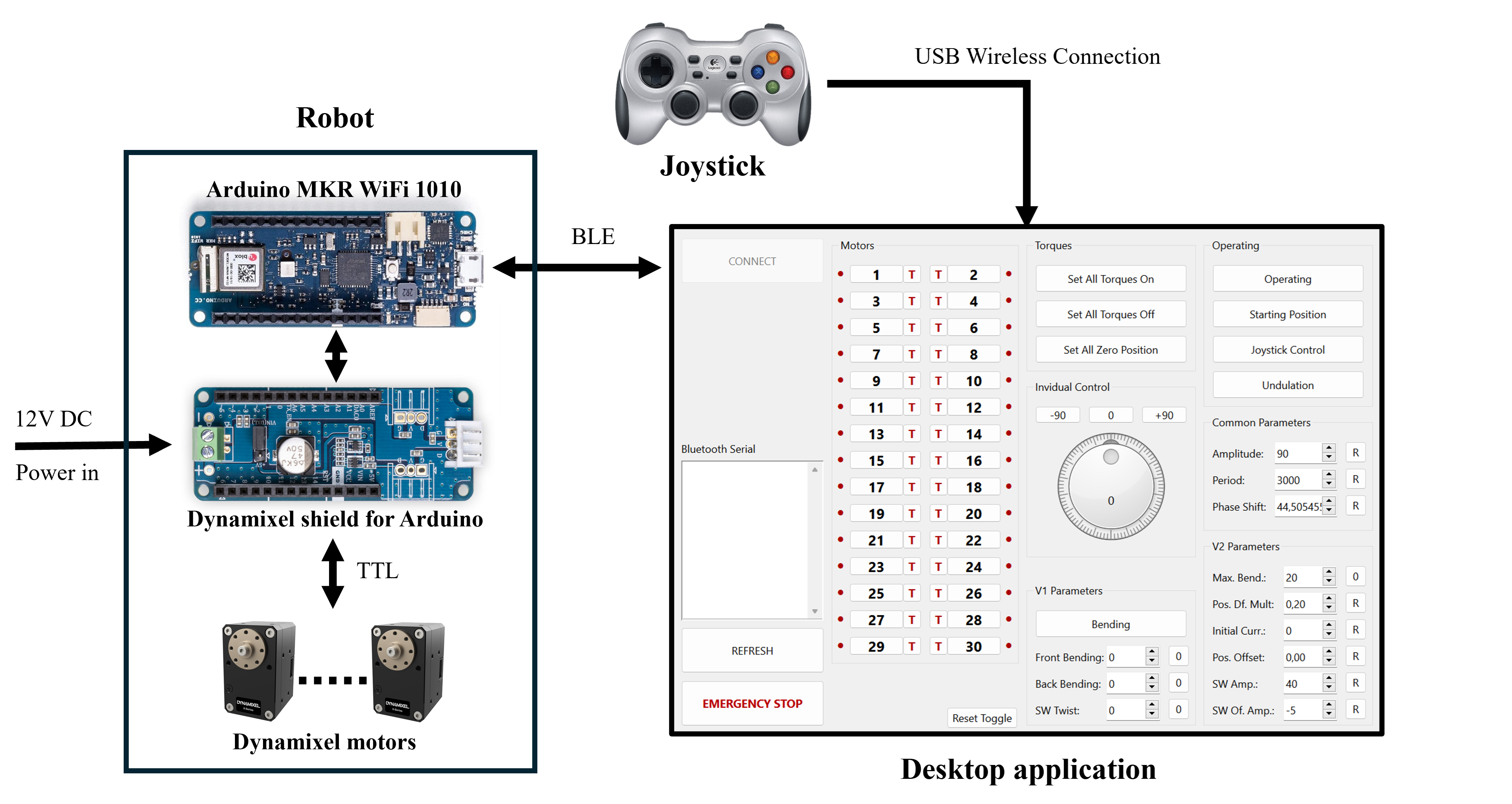}
\caption{Overall mechatronics system to actuate and teleoperate the robot: An Arduino board is connected to the motors through a Dynamixel shield. A desktop application was developed to communicate with the Arduino board and a joystick.
}
\end{figure}

\section{Locomotion Control}
\label{sec:locomotion_control}

To control the robot, we established a bending model for the continuum joints and implemented a motor position-control scheme to propagate flexural waves along the robot's length. Additionally, we provide a simple modification of the target angles in the motor control to achieve steerability for the robot.

\subsection{Bending Model of Continuum Joint}

The bending angle of our continuum joint is determined by the displacement of the pulling tendon. To establish the relationship between the motor angle and the joint's bending angle, we adopt the constant curvature assumption, as utilized in many other bending models of continuum mechanisms~\cite{webster2010design}.

\begin{figure}[t]
	\subfloat[]{
		\includegraphics[height=0.42\columnwidth]{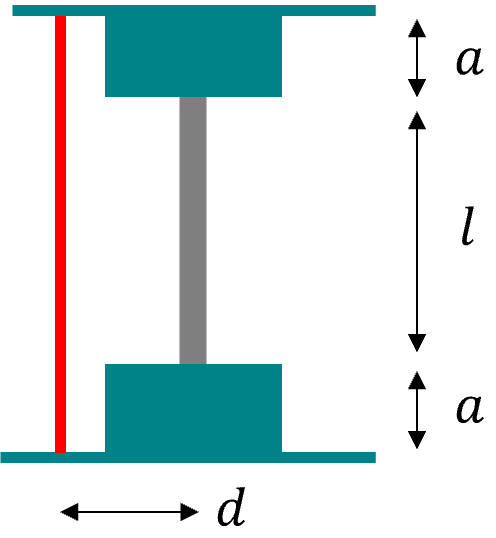}
	}
        ~
	\subfloat[]{
		\includegraphics[height=0.46\columnwidth]{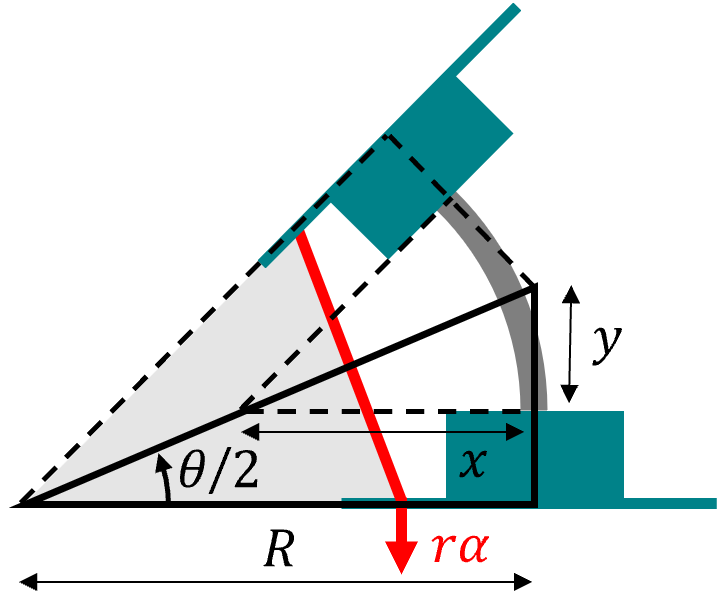}
	}
\caption{Bending geometry of continuum joints: (a) before bending, and (b) after bending. The values of $a,d,l$ and $r$ in our continuum joints are 7 mm, 10.5 mm, 20 mm, and 10.5 mm, respectively.
The red line represents the left-side tendon, while the right-side tendon is not visualized.}
\label{fig:bending_model}
\end{figure}

Let $\alpha \in \Re$ denote the motor angle and $\theta \in \Re$ denote the joint's bending angle, as defined in Fig.~\ref{fig:joint_actuation}(a). When the motor is rotated at a positive angle (i.e., counter-clockwise rotation), the left-side tendon is pulled by $r \alpha$, where $r \in \Re$ is the radius of the motor's rotor. This situation is illustrated in Fig.~\ref{fig:bending_model}. Note that the bending angle $\theta$ is identical to the central angle of the backbone arc.

Let us define a set of length variables, $a,d,l,x,y$ and $R$, as denoted in Fig.~\ref{fig:bending_model}.
From the black triangle in Fig.~\ref{fig:bending_model}(b), we obtain
\begin{eqnarray}
    R &=& (y+a) \cot \frac{\theta}{2}, \label{eqn:R}\\
    y &=& x \tan \frac{\theta}{2},
    \label{eqn:y}
\end{eqnarray}
where $x$ is the radius of the backbone arc, i.e., $x = l / \theta$.
By substituting this equation and (\ref{eqn:y}) into (\ref{eqn:R}), we can express $R$ as
\begin{equation}
    R = \frac{l}{\theta} + a \cot \frac{\theta}{2}.
    \label{eqn:R2}
\end{equation}
The length of the red line in Fig.~\ref{fig:bending_model}(b) is the retracted tendon length, which is $2a + l - r \alpha$. The isosceles triangle including the red line (i.e., the gray triangle in Fig.~\ref{fig:bending_model}(b)) can be divided in half into two right triangles, where we can derive
\begin{eqnarray}
    2a + l - r \alpha = 2 (R-d) \sin \frac{\theta}{2}.
    \label{eqn:red}
\end{eqnarray}
Finally, substituting (\ref{eqn:R2}) into (\ref{eqn:red}) yields the motor angle $\alpha$ expressed in terms of the bending angle $\theta$:
\begin{eqnarray}
    \alpha = \frac{2a}{r} \left(1 - \cos \frac{\theta}{2} \right) + \frac{l}{r}  \left(1 - \frac{2}{\theta} \sin \frac{\theta}{2} \right) + \frac{2d}{r}  \sin \frac{\theta}{2}.
    \label{eqn:bending_model}
\end{eqnarray}
When the right-side tendon is pulled by a negative $\alpha$, the above equation remains almost identical, with the signs of the first two terms on the right side being flipped.

Note that Equation (\ref{eqn:bending_model}) determines the motor angle $\alpha$ given a desired bending angle $\theta$. When the bending angle is smaller than $180^\circ$ (i.e., $0 < \theta < 180^\circ$), the derivative of (\ref{eqn:bending_model}) with respect to $\theta$ is always positive. This means that $\alpha$ monotonically increases with increasing $\theta$, indicating that smooth motor swings generate smooth joint swings without introducing any jittering.

\subsection{Motor Control}

The control of the snake robot is governed by the coordinated movement of its joints, which is crucial for achieving the desired locomotion patterns. In our robot, each motor connected to the joints is position-controlled by a sinusoidal function of time. 

As the joints must bend to propagate waves through the robot length, the sinusoidal function contains two terms varying terms, including a time-dependent term and a motor position-dependent term, as similarly presented in \cite{ha2023robotic}:
\begin{equation}
\alpha_i(t) = \alpha_\text{max} \sin(\phi_0(t) + \psi_i), \label{eqn:motor_pos}
\end{equation}
where $\alpha_i(t) \in \Re$ represents the motion of the $i$-th motor, $t \in \Re$ is the time variable, $\alpha_\text{max} \in \Re$ is the amplitude of motor swing, and $\phi_0(t) \in \Re$ and $\psi_i \in \Re$ are the time-dependent and motor position-dependent terms, respectively, defined as
\begin{eqnarray}
    \phi_0(t) &=& 360^\circ \frac{t}{T}, \\
    \psi_i &=& -i \Delta \psi.
\end{eqnarray}
Here $T \in \Re$ is the period of undulation, $i$ is the motor index, and $\Delta \psi$ is the phase shift between motors. Note that the amplitude $\alpha_\text{max}$ is computed using Equation (\ref{eqn:bending_model}) for the desired bending swing range.

\subsection{Steering Control}

Steering the direction of the robot's locomotion is an important functionality when the robot is teleoperated with a joystick. Implementing steerability in snake locomotion is not straightforward, particularly for sidewinding and backward locomotion. Meanwhile, forward locomotion can be relatively easily steered by curving the entire body in the desired direction. This can be achieved by adding an angular bias to each motion's control, as follows:
\begin{equation}
    \alpha_i(t) \leftarrow \alpha_i(t) + \alpha_\text{bias},
\end{equation}
where the bias angle $\alpha_\text{bias} \in \Re$ is positive for a left turn and negative for a right turn. We mapped this value linearly to the input of one of the analog sticks of our joystick, providing intuitive control when steering the robot.

\section{Experiments}
\label{sec:experiments}

We validated our snake robot through a series of indoor experiments. We primarily evaluated the locomotion capabilities of our robot, focusing on validating the effectiveness of global bending and twisting actuation in producing snake locomotion in various directions. The efficiency of the robot's movements was assessed by measuring locomotion velocities. Additionally, we demonstrated the robot's steerability by navigating a $90^\circ$ corner in an indoor corridor.

Detailed procedures for motion parameter selection and experiment conduction are provided first, followed by the assessment results of the robot's locomotion.

\subsection{Motion Parameter Selection}

The motion parameters of the motor swing include $\alpha_\text{max}$, $T$, and $\Delta \psi$, as used in (\ref{eqn:motor_pos}). Additionally, the locomotion parameters are the pulling displacements of the globally routed tendons. These tendon displacements for vertical upper bending, vertical lower locomotion, and axial twisting are denoted as $L_\text{u}$, $L_\text{l}$, and $L_\text{t}$, respectively. We remark again that the tendons routed through the back of the robot are pulled to create upper bending for forward locomotion, those on the abdomen are pulled to cause lower bending for backward locomotion, and the spiral tendons are pulled for sidewinding.

Some of these motion parameters were initially selected based on desired motion properties and subsequently fine-tuned. 
For example, the amplitude of the motor swing, $\alpha_\text{max}$, determines the curving shape of the robot. As we did in our previous work~\cite{ha2023robotic}, the robot's initial configuration for each locomotion mode was selected to consist of three curving sections, each making a $180^\circ$ turn, similar to the configurations in Figs.~\ref{fig:forward}, \ref{fig:backward}, and \ref{fig:side}. With $12$ joints in our robot, each curving section comprises four joints. Therefore, $\alpha_\text{max}$ was chosen to be $\alpha_\text{max} = 90.18^\circ$ for the first four joints to achieve a total bending angle of $180^\circ$ at $t=0$ sec.
This value was fine-tuned for each motor during experiments.
Similarly, the phase shift between motors, $\Delta \psi$, was chosen as $\Delta \psi = 45^\circ$ so that each curving section comprises four motors.

The time period of undulation, $T$, and the globally routed tendon displacements were determined empirically as given in Table~\ref{tab:parameters}. These parameters were chosen to create the most rapid motions without the robot to flip over. 
Additionally, we observed that the head part curves less than the tail part does, likely due to the dynamic effect of the undulatory motion. To compensate for this, we added a small angle to $\alpha_\text{max}$ that gradually decreased for $27.5^\circ$ to $0^\circ$ from the head to the tail.

\begin{table}[t!]
\label{tab:parameters}
\setlength{\tabcolsep}{8pt}
\renewcommand{\arraystretch}{1.5}
\begin{center}
\caption{Locomotion parameters}
\label{locomotionTable}
{\scriptsize
\begin{tabular}{ c | c | c | c }
\cline{2-4}
\multicolumn{1}{c}{} & \multicolumn{3}{c}{Locomotion type}\\
\cline{2-4}
\multicolumn{1}{c}{} & Forward & Backward & Sidewinding\\
\hline\hline
Amplitude, $\alpha_\text{max}$ ($^\circ$) & \multicolumn{3}{c}{$90.18$}\\
\hline
Period, $T$ (s) & \multicolumn{3}{c}{$3.0$}\\
\hline
Phase shift, $\Delta \psi$ ($^\circ$) & \multicolumn{3}{c}{$45$}\\
\hline
Upper Tendon Pulling, $L_{u}$ (mm) & $52.4$ & - & $14.0$ \\
\hline
Lower Tendon Pulling, $L_{l}$ (mm) & - & $69.8$ & -\\
\hline
Spiral Tendon Pulling, $L_{t}$ (mm) & - & - & $27.9$\\
\hline
\end{tabular}
}
\end{center}
\end{table}

\subsection{Experimental Procedure}

All parameters were set and managed in our desktop application, which was connected to the robot's Arduino board through BLE. A joystick was utilized to control the robot, with a button mapped to planar undulation and an analog stick used to steer the robot left or right. These commands were transmitted to the robot's Arduino board through the desktop application.

Experiments were conducted indoors on a tiled surface, with each tile measuring $30$ cm $\times$ $30$ cm. This setup facilitated our observation of the robot's movement and orientation. Upon establishing the BLE connection in each experiment, the robot was initialized in a starting position (i.e., a planar S-shape). The motor motion parameters were applied first, followed by the pulling displacements of the globally routed tendons.

We conducted two experiments to (i) validate the robot's ability to perform different types of snake locomotion and measure the performance, and (ii) demonstrate the robot's capability to steer in forward locomotion.

\subsection{Locomotion Capability and Performance Assessment}

\begin{figure}[t]
	\subfloat[]{
		\includegraphics[width=0.99\columnwidth]{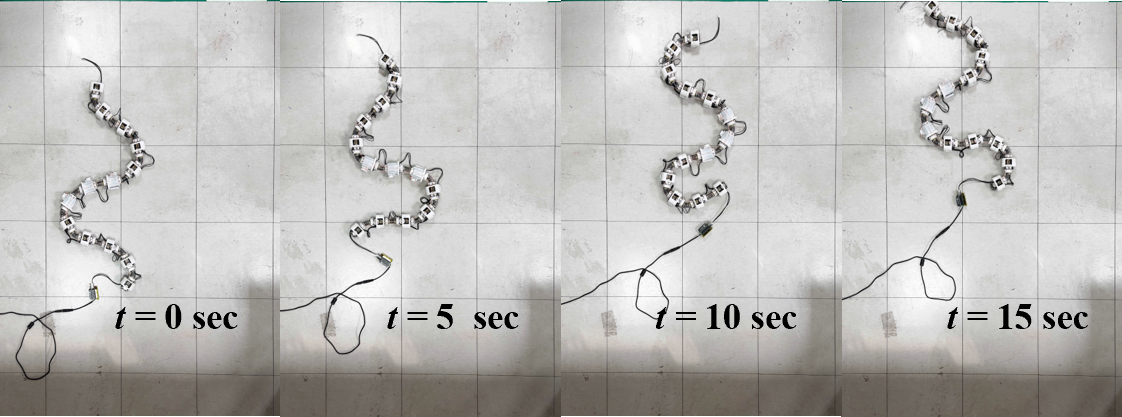}
	}
	\\
	\subfloat[]{
		\includegraphics[width=0.99\columnwidth]{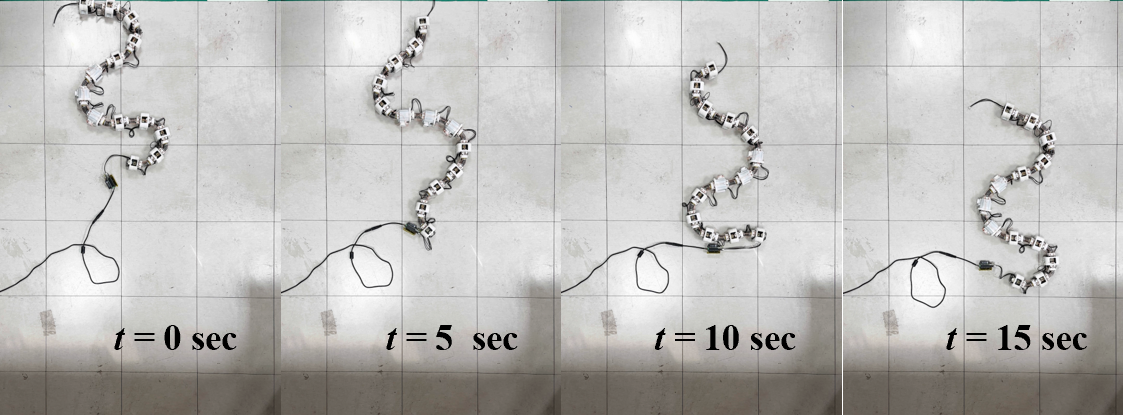}
	}
	\\
	\subfloat[]{
		\includegraphics[width=0.99\columnwidth]{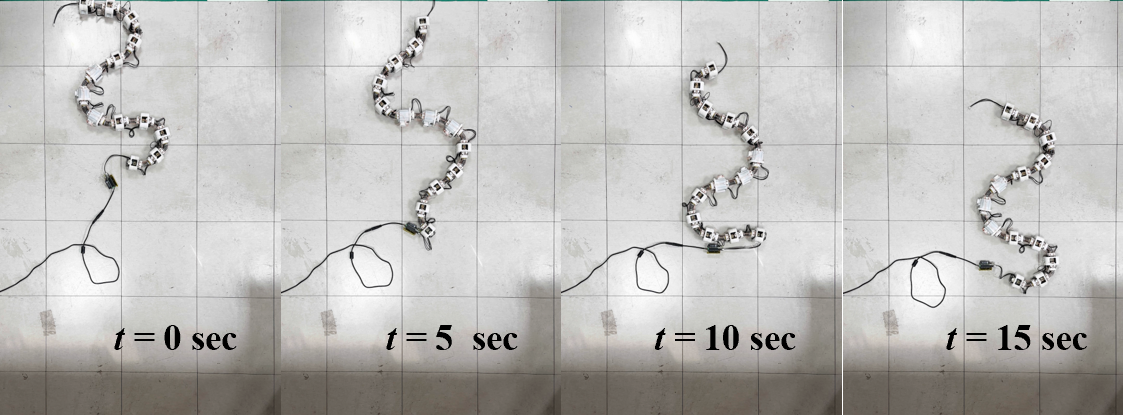}
	}
\caption{Various types of snake locomotion demonstrated by our robot: (a) forward locomotion, (b) backward locomotion, and (c) sidewinding.}
\label{fig:exp}
\end{figure}

The primary aim of these experiments was to verify if the robot could produce various directional snake movements by actuating the globally routed tendons. As shown in Fig.~\ref{fig:exp}, our experiments demonstrated the robot's ability to perform various types of locomotion, As expected, pulling the tendons for vertical upper and lower bending made the robot move forward and backward, respectively, and pulling the spiral tendons generated an axial twist, resulting in sidewinding motion.

Beyond demonstrating the locomotion capability, we measured the velocities of these motions as a performance metric. The resulting velocities are given in Table~\ref{tab:performances}. The forward, backward, and sidewinding motions achieved speeds of $27.6$ mm/sec, $35.5$ mm/sec, and $20.0$ mm/s, respectively

Although the robot demonstrated various locomotion patterns effectively, the achieved speeds were relatively slow. This is attributed to several factors. The tendon-driven mechanism and the use of compliant components, as opposed to rigid ones, introduced complexity and variability in the motion. The primary issue we experienced in the experiments was the robot flipping over, causing signal wires, tendons, and springs to tangle together. The cause of this behavior was identified as torsional backlashes and insufficient torsional stiffness of the continuum joints. Considering that these joints were built by hand using promptly available components in the market, better manufacturing may take a significant amount of time.

This behavior was more pronounced when the motion parameters were chosen for faster motion. Increasing the undulation frequency or actuating tendons for more vertical bending and axial twisting led to more instability in the motion.

\begin{table}[t!]
\label{tab:performances}
\setlength{\tabcolsep}{8pt}
\renewcommand{\arraystretch}{1.5}
\begin{center}
\caption{Performance measurements
}
\label{locomotionTable}
\begin{tabular}{ c | c | c | c }
\cline{2-4}
\multicolumn{1}{c}{} & \multicolumn{3}{c}{Locomotion type}\\
\cline{2-4}
\multicolumn{1}{c}{} & Forward & Backward & Sidewinding\\
\hline\hline
Speed (mm/s) & 27.6 & 35.5 & 20.0 \\
\hline
\end{tabular}
\end{center}
\end{table}

\subsection{Steerability Demonstration}

Lastly, we demonstrated the steerability of our robot using joystick inputs. We mapped an analog stick to a continuous steering input that adds a bias angle to the motor positions. We tested our robot by teleoperating it to navigate a $90^\circ$ corner in an indoor corridor and confirmed that the robot could smoothly turn around the corner. This motion is shown in Fig.~\ref{fig:steering} and can also be found in the supplementary video.

\begin{figure}[t!]
    \centering
    \includegraphics[width=3.5in]{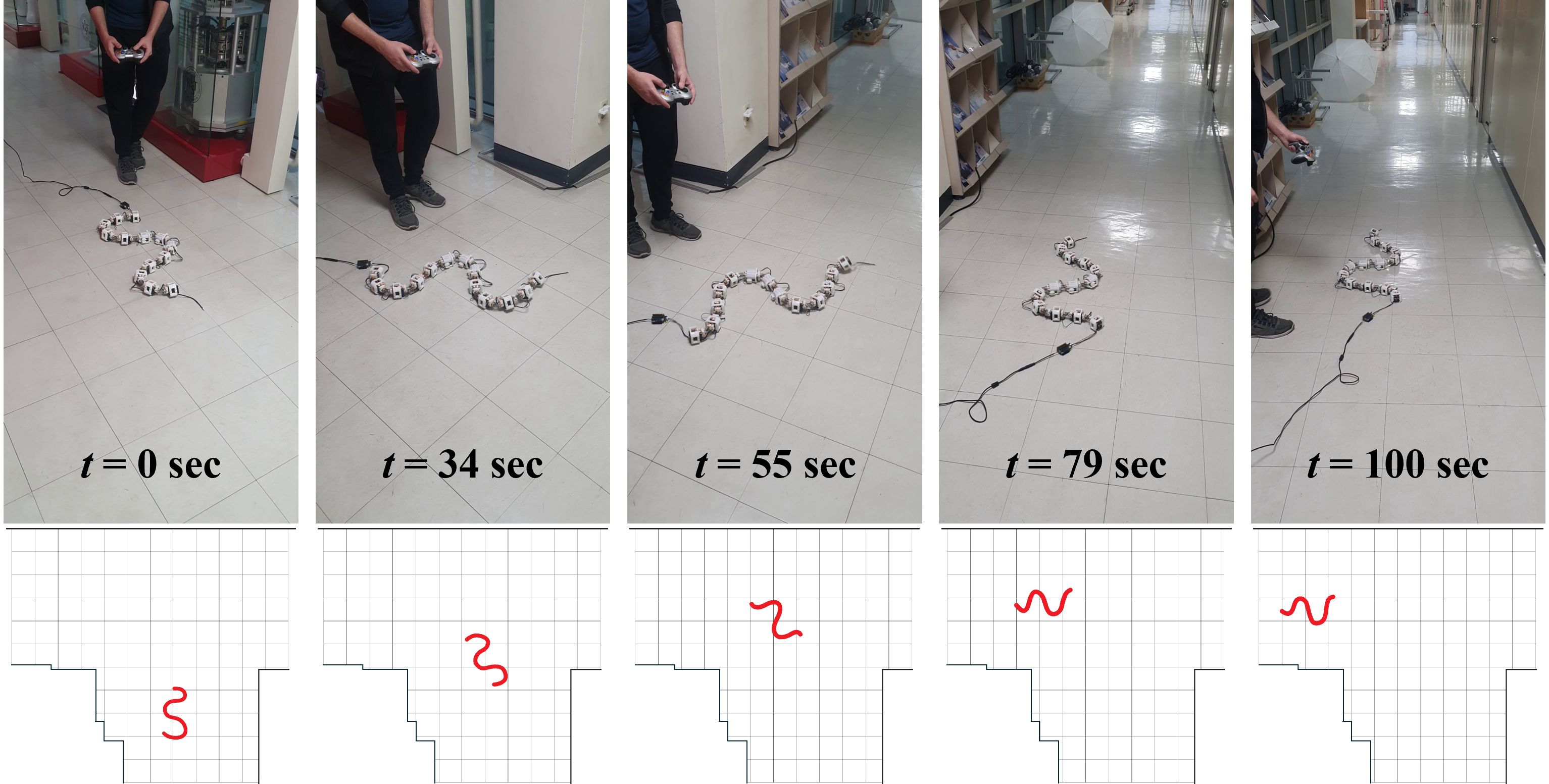}
    \caption{Robot's steerability demonstrated by navigating a $90^\circ$ corner in an indoor corridor. Each tile block on the floor is $30$ cm $\times$ $30$ cm.}
    \label{fig:steering}
\end{figure}

\section{Conclusion}
\label{sec:conclusion}

In this paper, we presented the development and validation of a tendon-driven compliant snake robot capable of global bending and twisting actuation. Building on our previous work on snake locomotion principles, we designed a robotic mechanism that emulates the natural ground contact of biological snakes, utilizing body compliance and gravitational deflection. The body compliance was achieved through compliant continuum joints used to concatenate modules. A desktop application with a wireless joystick interface was developed to enable intuitive teleoperation.

Through a series of indoor experiments, we validated the robot's locomotion capabilities, demonstrating effective forward, backward, and sidewinding movements. The measured velocities of these motions were $27.6$ mm/sec for forward, $35.5$ mm/sec for backward, and $20.0$ mm/sec for sidewinding. Moreover, we introduced a method to steer the robot by adding bias angles to the motor control and showcased the robot's steerability by successfully navigating a $90^\circ$ corner.

The achieved speeds were relatively slow to prevent the robot from flipping over, which was caused by factors such as torsional backlashes and insufficient torsional stiffness of the continuum joints. These issues, primarily attributed to the handmade joints built with readily available components, suggest that improved manufacturing techniques are required to enhance performance. Therefore, our future work will focus on refining the robot's design and control to increase its speed and stability. This includes enhancing the torsional stiffness of the joints and optimizing the locomotion parameters for more efficient locomotion.

\bibliography{ref}
\bibliographystyle{ieeetr}



\end{document}